# Adaptive Dual-Weighted Gravitational Point Cloud Denoising Method

Ge Zhang[1], Chunyang Wang[2*], Bin Liu[1], Guan Xi[1]

[1] School of Optoelectronic Engineering, Xi'an University of Technology, Shaanxi, Xi'an, 710000;
[2] Xi'an University of Technology, Xi'an Key Laboratory of Active Optoelectronic Imaging Detection Technology, Shaanxi, Xi'an, 710021;

**Abstract** High-quality point cloud data is a critical foundation for tasks such as autonomous driving and 3D reconstruction. However, LiDAR-based point cloud acquisition is often affected by various disturbances, resulting in a large number of noise points that degrade the accuracy of subsequent point cloud object detection and recognition. Moreover, existing point cloud denoising methods typically sacrifice computational efficiency in pursuit of higher denoising accuracy, or, conversely, improve processing speed at the expense of preserving object boundaries and fine structural details, making it difficult to simultaneously achieve high denoising accuracy, strong edge preservation, and real-time performance. To address these limitations, this paper proposes an adaptive dual-weight gravitational-based point cloud denoising method. First, an octree is employed to perform spatial partitioning of the global point cloud, enabling parallel acceleration. Then, within each leaf node, adaptive voxel-based occupancy statistics and k-nearest neighbor (kNN) density estimation are applied to rapidly remove clearly isolated and low-density noise points, thereby reducing the effective candidate set. Finally, a gravitational scoring function that combines density weights with adaptive distance weights is constructed to finely distinguish noise points from object points. Experiments conducted on the Stanford 3D Scanning Repository, the Canadian Adverse Driving Conditions (CADC) dataset, and in-house RUBY PLUS LiDAR point clouds acquired in our laboratory demonstrate that, compared with existing methods, the proposed approach achieves consistent improvements in F1, PSNR, and Chamfer Distance (CD) across various noise conditions while reducing the single-frame processing time, thereby validating its high accuracy, robustness, and real-time performance in multi-noise scenarios.

**Keyword** Lidar; Point cloud denoising; Adaptive dual-weight gravitational model; Octree-based spatial partitioning

**Classification number of the middle image** O436    **Document code** A

## 1 Introduction

Point cloud data constitutes a discrete set of spatial points that encapsulate the three-dimensional information of scanned objects, providing an accurate and comprehensive reflection of their dimensions and morphology [1]. As a fundamental form of three-dimensional representation, the point cloud data format is distinguished by its simplicity and broad data acquisition sources, leading to its widespread adoption across various fields, including robotics [2], autonomous driving [3], and 3D reconstruction [4]. However, during the actual data acquisition process, various factors such as sensor hardware inaccuracies, surface reflectance properties, and external environmental disturbances introduce noise into the point cloud data. Different types of acquisition devices contribute varying degrees and forms of such noise. The objective of point cloud denoising is to preserve the authentic geometric structures of the object while optimally reconstructing a clean and precise three-dimensional shape from the raw data. Therefore, denoising is typically regarded as an essential preprocessing step in point cloud processing workflows, providing a high-quality data foundation for subsequent tasks such as detection, classification, and tracking.

Currently, scholars both domestically and internationally have primarily proposed learning-based and non-learning-based denoising methods to address the issue of noise in point clouds [5]. Learning-based denoising methods primarily encompass PointCleanNet [6], PointFilter [7], GDPNet [8], and PDFlow [9]. These methods design and train different network architectures to predict displacement vectors on object surfaces, with the ability to capture object shape features from noisy point clouds. However, learning-based approaches often require a substantial amount of object point cloud data for training neural network models, resulting in significant computational resource consumption and extended processing times. Furthermore, learning-based denoising techniques are limited to specific scenarios or object point cloud datasets, lacking universal applicability. Therefore, this paper proposes a non-learning-based point cloud denoising method that accomplishes the denoising task simply and efficiently. Non-learning-based point cloud denoising methods primarily include filter-based approaches and optimization-based techniques. The filter-based approaches include bilateral filtering [10] and guided filtering methods [11-14]. Optimization-based approaches primarily include the Moving Least Squares (MLS) method [15] and denoising techniques based on the

Locally Optimal Projection (LOP) framework [16]. Non-learning approaches primarily encounter three key challenges. Firstly, during the denoising process, valuable data may be eliminated along with the noise. Secondly, insufficient denoising can result in blurred object edge feature information, while excessive denoising may lead to the loss of critical edge features. Thirdly, there is the issue of poor real-time performance in noise reduction.

To address the issues outlined above, I propose an adaptive dual-weight gravitational point cloud denoising method that enables rapid and precise noise removal from point cloud data. Section 2 primarily reviews the current state of research on relevant denoising algorithms. Section 3 presents the fundamental theory of the proposed method by providing the formula for adaptive dual-weight gravitational calculation. Section 4 presents the experimental validation. By conducting simulation experiments on both public and proprietary datasets and performing comparative analyses with other point cloud denoising techniques across key technical metrics, we validated the effectiveness of the proposed method. Section 5 presents the conclusions and future directions.

The main contributions of this paper are summarized as：

(1) This paper proposes an adaptive dual-weight gravitational point cloud denoising method. An octree-based blockwise parallel local processing framework is constructed, within which each leaf node sequentially performs adaptive voxel outlier removal, kNN low-density point removal, and computation of a density–distance dual-weight gravitational score, thereby achieving fine-grained denoising of point clouds.

(2) This paper designs a noise-point removal mechanism that balances accuracy and efficiency. First, voxel occupancy counts are used to remove outlier noise points at the spatial scale, and then kNN density is employed to further eliminate low-density noise points, effectively compressing the candidate set. On this basis, weighted gravitational scores are computed only for a small number of candidate points, which significantly reduces the overall computational cost. Structural ablation results demonstrate that adaptive voxel-based outlier removal and kNN-based low-density point removal can substantially reduce the computational load and provide an efficient and reliable candidate set for subsequent weighted gravitational score computation.

(3) This paper proposes a density–distance dual-weight gravitational scoring function to achieve fine-grained discrimination of noise points. Building on the gravitational-feature-based point cloud denoising model, kNN density weights and adaptive distance weights are introduced to reweight the pairwise gravitational interactions between points, thereby constructing a gravitational scoring function that jointly accounts for local density and spatial configuration and is used to distinguish object points from residual noise. Experimental results demonstrate that the dual-weight gravitational scoring function outperforms comparative methods in preserving edge shape characteristics.

## 2 Related Work

### 2.1 Current State of Research

Currently, both domestic and international scholars have proposed a variety of denoising algorithms to address the issue of noise in point clouds. Han et al. [17] proposed a denoising method for point cloud data based on guided filtering. However, this approach has shown suboptimal performance when dealing with densely distributed noise points. Zhao et al. [18] proposed a classification-based point cloud denoising algorithm, which utilizes different filtering techniques to process the point cloud according to varying noise scales. Peng et al. [19] proposed a point cloud denoising algorithm based on least squares density estimation. The method first constructs a high-dimensional feature density space, then employs the least squares method to solve for density fitting curves across each dimension, and finally extracts the intersection of valid point sets from each dimension according to a standard thresholding criterion. This algorithm demonstrates high accuracy and effective outlier removal for point clouds in urban environments; however, its high dimensionality results in substantial computational complexity and extended processing times. Wu et al. [20] proposed a neighborhood-adaptive filtering algorithm for point clouds. First, to address the shrinkage issue in filtering at the original and mean points, I propose a hybrid super-sampling strategy. Then, I employ an adaptive neighborhood selection approach to maintain the filtering accuracy of key features. Finally, I perform iterative refinement along the gradient direction and use maximum likelihood estimation to obtain the optimal filtering result. This method results in significant errors when processing highly noisy point clouds, thereby failing to achieve satisfactory filtering performance and precision. Huang et al. [21] proposed a segmented denoising method for point clouds based on adaptive thresholding. Based on the Euclidean distance between noise points and non-noise points, noise points are categorized into distant-signal and proximate-signal noise. Subsequently, a threshold-adaptive denoising algorithm based on nonlinear functions is applied to the distant-signal noise, while a curvature-based denoising algorithm is employed for the proximate-signal noise. This algorithm demonstrates high accuracy; however, it involves a considerable number of tunable parameters. Long et al. [22] proposed a point cloud denoising method based on image segmentation techniques. A two-dimensional point cloud mapping image is constructed based on the three-dimensional point cloud reconstructed from structured light fringe projection. Image segmentation of the two-dimensional point cloud mapping is performed using image thresholding and region growing techniques, and the segmented noise regions are identified and eliminated. Finally, a new three-dimensional reconstruction is performed on the updated two-dimensional point cloud projection image to obtain a denoised point cloud. This method circumvents the complex computation inherent to three-dimensional data; however, it is applicable solely to structured-light point clouds. Luo et al. [23] proposed a point cloud denoising algorithm that integrates enhanced radius filtering with local plane fitting. To precisely eliminate noise points, I categorize them into distant and proximal noise points based on their Euclidean distances from the object point cloud, then employ tailored denoising strategies for each category in succession. S et al. [24] proposed a point cloud guided filtering algorithm based on optimally weighted neighborhood features to address small-scale noise points. The selection of the optimal neighborhood is achieved using an entropy-based information function. Feature point identification is achieved by integrating surface curvature, normal variation, and distance-based feature descriptors. I implemented adaptive neighborhood growth around feature points and employed a guided filtering algorithm to achieve anisotropic smoothing for both feature and non-feature regions of complex surface components. This method demonstrates a more pronounced smoothing effect and superior performance in feature preservation; however, it is computationally intensive and requires longer processing times. Shi et al. [25] proposed a point cloud denoising algorithm based on a gravitational feature function. Building upon the universal law of gravitation, this algorithm derives a novel gravitational formula specifically tailored for point clouds. By calculating the gravitational threshold according to the defined formula, point cloud noise can be effectively filtered. This method is capable of removing both sparse and dense drift noise; however, it is computationally intensive and time-consuming to process large datasets. Since the bilateral filtering denoising algorithm relies on point normals, the accuracy of these normals significantly affects the algorithm's performance. Yang et al. [26] proposed a novel robust bilateral filtering method for point cloud denoising. First, principal component analysis is employed to estimate the initial normals of the point cloud. Subsequently, spherical fitting is performed to classify the points into edge and planar categories. Iterative weighted principal component analysis is then applied to refine the normals of edge points. Finally, the enhanced normals are integrated with bilateral filtering to achieve effective point cloud denoising. This method has primarily been validated under simple Gaussian noise conditions; further research is needed to assess its adaptability to complex environmental noise and real-world LiDAR scenarios.

In summary, most existing point cloud denoising methods primarily rely on the local neighborhood information of the point cloud, lacking comprehensive global modeling of the overall point cloud structure. This limitation compromises their effectiveness in addressing noise points located near the object surface, as well as cluster-distributed outlier noise. As a result, it becomes challenging to balance noise suppression and the preservation of genuine data points, ultimately restricting the overall denoising performance. The denoising method based on the gravitational characteristic function demonstrates superior robustness in eliminating both sparse and dense drift noise. However, its computational complexity is relatively high, resulting in insufficient efficiency when processing large-scale point clouds. To address the aforementioned issues, this paper proposes an adaptive dual-weighted gravitational point cloud denoising method, which further enhances denoising accuracy and improves computational efficiency.

## 2.2 Principle of the Gravitational Characteristic Function Denoising Method

This section presents the fundamental theory of the point cloud denoising method based on the gravitational feature function, thereby laying the theoretical groundwork for the derivations discussed in Section 3. This method first calculates the centroid of the point cloud based on its spatial distribution, representing the location of the average mass distribution. Subsequently, the local spherical neighborhood of the centroid is determined according to both the number of points and their spatial arrangement within the point cloud. Based on the gravitational computation formula for point clouds, each point's local spherical neighborhood is regarded as a homogeneous sphere. The gravitational force between each homogeneous sphere and the centroid is calculated, and the resulting set of all gravitational values constitutes a gravitational feature function. Finally, noise within the point cloud is eliminated by applying the threshold of the gravitational feature reference function.

First, derive the method for calculating the centroid of the point cloud. Let the three-dimensional point cloud data be denoted as $P=\{p_1,p_2,...,p_N\}$. Among these $p_i=\{x_i,y_2,z_i\}$, N represents the total number of points within the point cloud dataset. $p_i$ denotes the three-dimensional coordinates of the $i$-th point. Subsequently, I calculated the centroid of the entire point cloud, denoted as $\theta$. The calculation formula is as follows:

$$\theta = \frac{1}{N}\sum_{i=1}^{N}p_i = (\frac{1}{N}\sum_{i=1}^{N}x_i, \frac{1}{N}\sum_{i=1}^{N}y_i, \frac{1}{N}\sum_{i=1}^{N}z_i,) \quad (1)$$

This point represents the mean spatial position of the single-frame point cloud and serves as the reference for subsequent gravitational calculations.

To estimate the spatial density of the point cloud, first determine the boundary limits of the point cloud along each axis.

$$\begin{aligned} H_x &= \max_i(x_i)-\min_i(x_i) \\ H_y &= \max_i(y_i)-\min_i(y_i) \\ H_z &= \max_i(z_i)-\min_i(z_i) \end{aligned} \quad (2)$$

Among them, $H_x, H_y$, and $H_z$ represent the extents of the point cloud data along the $x$, $y$, and $z$ axes, respectively. Accordingly, a local neighborhood search radius $R$ is defined for subsequent neighborhood search procedures.

$$R = \frac{6(H_xH_y + H_xH_z + H_yH_z)}{N} \quad (3)$$

In Equation (3), the numerator represents the surface area covered by the point cloud within the three-dimensional space, while the denominator indicates the total number of points in the point cloud dataset. Equation (3) effectively establishes a suitable neighborhood search radius by integrating constraints based on both the spatial extent of the point cloud and the scale of the point set.

For each point $p_i$, I calculate the Euclidean distance $d_i$ between it and the centroid of the point cloud.

$$d_i = \|p_i-\theta\| = \sqrt{(x_i-\theta_x)^2+(y_i-\theta_y)^2+(z_i-\theta_z)^2} \quad (4)$$

The set of neighboring points $N_i$ within a radius $R$ of the given point is defined as follows:

$$N_i = \{p_i \in P \mid \|p_j - p_i\| \leq R\} \quad (5)$$

Let $n_i = |N_i|$ denote the number of neighboring points. Based on the law of universal gravitation, the gravitational force $F_i$ at this point is defined as:

$$F_i = \frac{G \cdot n_i}{d_i^2} \quad (6)$$

$G$ represents the gravitational constant, which is used to adjust the magnitude. In this method, its value is directly assigned as $G = 6.67 \times 10^{-11}$. Equation (6) indicates that the structural significance of a given point is jointly determined by the "number of neighboring points" and the "distance from the center." A point is considered an "important point" if it is situated close to the center and possesses a higher number of neighbors.

To achieve automatic noise removal, an adaptive threshold $T$ is introduced, calculated with reference to the average density of the point cloud.

$$T = \alpha \cdot G \cdot \frac{H_x + H_y + H_z}{N} \quad (7)$$

$\alpha$ represents the experiential weighting parameter, which is set to 600 in this study to amplify the threshold magnitude. The criteria for determining the retention points are defined as follows:

$$F_i \geq \frac{T}{d_i} \quad (8)$$

Equation (8) takes into account both the distance $d_i$ from the point to the centroid and its gravitational strength $F_i$. A higher gravitational value $F_i$ indicates that the point is located within a dense region or near the structural center; conversely, lower values are more likely to represent noise.

The final denoised point set is as follows:

$$p_{denoised} = \left\{ p_i \in P \middle| F_i \geq \frac{T}{d_i} \right\} \quad (9)$$

## 3 Principle of the Adaptive Dual-Weighted Gravitational Denoising Method

This section presents the overall workflow and theoretical derivation of the proposed adaptive dual-weight gravitational point cloud denoising method. As illustrated in Fig. 1, the input point cloud is first recursively partitioned in space using an octree structure, decomposing the large-scale point cloud into multiple sub-leaf nodes. Parallel processing is then performed at the sub-leaf level, which significantly reduces the number of points processed in each neighborhood search and greatly enhances overall computational efficiency. Subsequently, within each sub-leaf, I adaptively estimate the average point spacing based on the size of the local bounding box and the number of points. Using this information, I generate a voxel grid and count the occupancy of each voxel. Voxels containing fewer points than the threshold are classified as outlier voxels and are removed, thereby effectively eliminating significant outlier points. Since the voxel size is automatically adjusted according to the distribution of point clouds within each lobe, this process exhibits an adaptive characteristic. In Fig. 1, the red circles indicate clearly isolated noise points that can be directly removed at this stage, while the blue circles denote suspicious noise points that cannot yet be definitively identified and thus require further screening in subsequent steps. Next, a k-nearest neighbor(kNN) search is performed on the voxel-filtered point set to calculate local point density and remove low-density points, thereby further compressing the candidate set while maximally preserving object edge points. Finally, a dual-weighted attraction scoring model is introduced into the refined candidate point set, incorporating both density weighting and adaptive distance weighting. Attraction scores are calculated for each point, and high-scoring points are retained based on a top-percentage selection strategy. This approach effectively removes noise points while preserving the geometric structure of the object's surface and edges.

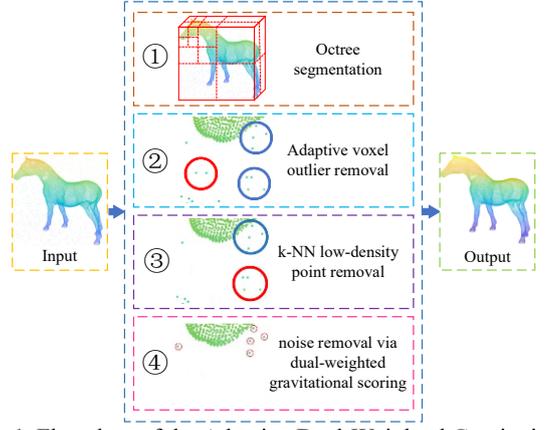

Fig. 1. Flowchart of the Adaptive Dual-Weighted Gravitational Denoising Method. In the diagram, points within blue circles indicate noise points that cannot be removed at the current step, while points within red circles represent noise points that can be eliminated at this stage.

An octree is an efficient three-dimensional spatial data structure that enables rapid space partitioning and nearest neighbor search, offering significant advantages in the processing and accelerated computation of point clouds. This paper first utilizes an octree to segment the point cloud, partitioning the point cloud into multiple sub-leaves. This approach enables parallel processing, thereby significantly enhancing computational efficiency. Let the point cloud be denoted as $P = \{p_1, p_2, ..., p_N\}$. The set of sub-leaves obtained after octree partitioning is denoted as $\{P_t\} t = 1, 2...n$. The value of $n$ is determined by the maximum number of points permitted within each leaf node and the minimum side length threshold of each leaf node in the octree algorithm. Each cotyledon contains $N_t = |P_t|$ points. According to Equation (2), let $(H_{tx}, H_{ty}, H_{tz})$ denote the side length of the bounding box enclosing the point cloud within the sub-leaf. Therefore, the volume of the point cloud contained in the sub-leaf can be expressed as:

$$V_t = H_{tx} \cdot H_{ty} \cdot H_{tz} \quad (10)$$

The denoising method for point clouds based on gravitational characteristic functions exhibits relatively high computational complexity, primarily because a substantial number of distinctly isolated noise points are also included in the pointwise gravitational calculations. The gravitational score of such isolated noise points is typically quite low, and these points are ultimately eliminated. However, their involvement in the calculation results in unnecessary computational overhead. Therefore, this paper introduces a pre-screening strategy before calculating the gravitational scoring. Priority should be given to identifying and removing clear outliers, isolated points, and low-density data points to ensure that they are excluded from subsequent weighted gravitational score calculations. This approach significantly reduces the number of data points involved in the computation, thereby decreasing computational complexity and enhancing processing speed.

First, adaptive voxelization is applied to remove outliers and isolated points. The adaptive voxel size $h$ is

$$h = \beta \left( \frac{V_t}{N_t} \right)^{1/3} \quad (11)$$

Since $V_t$ and $N_t$ represent the volume and number of contained points within the sub-leaf point cloud, respectively,

the voxel size is adaptively determined. $\beta$ represents the voxel enlargement coefficient. As $\beta$ increases, the voxels become larger, resulting in the removal of more outliers; however, this may also lead to the loss of crucial edge details of the object. A smaller $\beta$ result in smaller voxels, leading to a more conservative removal of outliers and more complete edges; however, this also increases the likelihood that noise points may be included in subsequent computations. For each voxel, I calculated the number of occupied points, $n(v)$, and retained only those voxels where the occupancy exceeded a defined threshold.

$$P_1 = \{p \in P_i | n(voxel(p)) \geq n_v\} \quad (12)$$

$n_v$ represents the minimum voxel occupancy. By removing outliers according to the formula above, we obtain a point set $P_1$.

To achieve robust and efficient low-density point removal, this study employs a kNN density estimation method based on the distance to the *k*-th nearest neighbor, denoted as $r_{K,i}$. This estimation utilizes an adaptive local scale determined by a parameter $r_{K,i}$, eliminating the need to specify a global radius. As a result, it demonstrates robust performance in scenarios involving non-uniform sampling and multiple scales. Compared with fixed-radius counting, kNN enables comparability across leaf nodes through the use of quantile-based thresholds. Each point consistently receives k neighbors, which prevents the discretization of counts at small radii and over-smoothing at larger radii, thereby ensuring greater control over estimation variance. Compared to kernel density estimation (KDE), kNN does not depend on the bandwidth parameter, which is often challenging to optimize, and can instead utilize $r_{K,i}$ directly as an adaptive bandwidth for subsequent distance weighting. Its computational and memory overheads are limited to a fixed-size $N \times K$ adjacency matrix, which facilitates vectorization and parallel processing. Compared to voxel counting, kNN offers point-level adaptability and is not constrained by grid resolution. Voxels are more suitable for an initial, rapid, and broad screening step, while kNN density estimation offers a second layer of fine-grained, adjustable statistical discrimination. In summary, kNN density estimation demonstrates clear advantages in parameter robustness, cross-scenario comparability, statistical stability, and engineering efficiency, and it synergizes effectively with the dual-weight gravitational scoring framework proposed in this study.

For the candidate set $P_1$, obtained after adaptive voxel-based outlier removal, I calculated the local density of each point based on the kNN. By filtering out low-density points using a density threshold, I reduce the computational load of subsequent gravitational calculations. For each set point $P_1$, retrieve its *k* nearest neighbors. Let the set of neighboring indices be denoted as $N_K(i)$. The Euclidean distance between the points is denoted as $d_{i,j} = \|p_i - p_j\|, j \in N_K(i)$. Define the distance to the *k* nearest neighbor as $r_{K,i} = \max_{j \in N_K(i)} d_{ij}$. Define a spherical region with radius $r_{K,i}$ as the minimal voxel encompassing the *k* nearest neighbor points. Treat each point in the point cloud as a mass point with unit mass; the density at point $p_i$ is defined as:

$$\rho_i = \frac{K}{\frac{4}{3}\pi r_{K,i}^3} \quad (13)$$

Calculate the density values for $p_i$, set a threshold at $q\%$, and retain only those points with densities equal to or exceeding this threshold.

$$P_2 = \{p_i \in P_1 | \rho_i \geq prctile(\rho, q)\} \quad (14)$$

After applying adaptive voxel-based outlier removal and kNN low-density point filtering, isolated and sparse noise points have been effectively eliminated. As a result, these points are excluded from subsequent weighted attraction score calculations, significantly reducing unnecessary computational overhead. Next, I calculate the weighted attraction score $F_i$ for each point *i* in the candidate set $P_2$. $F_i$ consists of three components: density weighting $w_{den}(i,j)$, adaptive distance weighting $w_{dis}(i,j)$, and gravitational kernel $k(d_{ij})$. The weighted attraction score $F_i$ for point *i* is obtained by sequentially summing the product of the three values for each neighbor *j* among the k nearest neighbors $N_K(i)$ of point *i*. This approach simultaneously emphasizes the contributions of both "high-density" and "closer-proximity" neighboring regions, effectively highlighting the object structural points while attenuating the interference from distant or low-density points. The calculation is detailed in Equation (15):

$$F_i = \sum_{j \in N_K(i)} w_{den}(i,j) \cdot w_{dis}(i,j) \cdot k(d_{ij}) \quad (15)$$

(1) density weighting

Because the overall sampling intensity varies among different leaves, directly using $\rho$ would introduce scale bias; therefore, we employ median normalization to achieve cross-leaf alignment.

$$\rho_{med} = median\{\rho_i\}, i = 1, 2...n \quad (16)$$

The density weight between point *i* in the candidate set $P_2$ and point *j* among its kNN is defined as follows:

$$w_{den}(i,j) = \left(\frac{\rho_i \rho_j}{\rho_{med}^2 + \varepsilon}\right)^{\alpha/2} \quad (17)$$

In equation (17), $\varepsilon$ is a constant introduced to prevent instability caused by the denominator approaching zero in cases of extreme sparsity; therefore, $\varepsilon$ should be assigned a very small positive value. $\alpha$ represents the density weighting index, which determines the strength of density weighting. As $\alpha$ increases, the system becomes more effective at suppressing sparse noise; however, excessively large values may lead to over-denoising.

(2) distance weighting

For the candidate point $p_i$, the adaptive bandwidth $\sigma_i$ is defined as follows:

$$\sigma_i = \sigma \cdot r_{K,i} \quad (18)$$

In Equation (18), $\sigma > 0$ serves as the global scaling factor. This configuration is designed to allow the bandwidth of distance weighting to adapt automatically to the local sampling scale, thereby achieving consistent smoothing and discriminative capability even under non-uniform density conditions. In regions of dense point clouds, the parameter $r_{K,i}$ is small, resulting in a smaller value for $\sigma_i$. This leads to a narrower distance weighting, effectively reducing excessive smoothing and aiding in the preservation of edge details. In sparse regions of the point cloud, the parameter $r_{K,i}$ is large, which results in a higher value of $\sigma_i$ and helps mitigate the issue of insufficient local information.

$$w_{dis}(i,j) = \exp(-\frac{d_{ij}^2}{2\sigma_i^2}), j \in N_K(i) \quad (19)$$

In Equation (19), $\exp\left(-d_{ij}^2/\left(2\sigma_i^2\right)\right)$ represents a smoothly varying soft cutoff, where contributions from nearby neighbors are significant, while those from distant neighbors decay exponentially.

(3) Gravitational core

Inspired by universal gravitation $F \propto 1/a^2$, which follows an inverse square law attenuation, we introduce the gravitational kernel as follows:

$$k(d_{ij}) = \frac{1}{d_{ij}^2 + \varepsilon} \quad (20)$$

$\varepsilon$ is assigned an infinitesimal positive value to ensure numerical stability as $d_{ij}$ approaches zero.

In summary, by substituting Equations (17), (19), and (20) into Equations (15), the weighted gravitational score $F_i$ for point $i$ is obtained as follows:

$$F_i = \sum_{j \in N_K(i)} \left(\frac{\rho_i \rho_j}{\rho_{med}^2 + \varepsilon}\right)^{\alpha/2} \cdot \exp\left(-\frac{d_{ij}^2}{2(\sigma r_{K,i})^2}\right) \cdot \frac{1}{d_{ij}^2 + \varepsilon} \quad (21)$$

In the computation of weighted gravitational values, density weights are constructed using kNN density estimates and normalized by the median. This method amplifies the contribution only when both points reside in high-density structural regions, while automatically reducing the weight for sparse or outlier point pairs, thereby enhancing the signal-to-noise ratio of structural points. The distance weights are determined using a Gaussian kernel with bandwidth $\sigma_i = \sigma \cdot r_{K,i}$, enabling an adaptive soft cutoff in accordance with the local sampling scale: in densely sampled regions, the bandwidth is narrowed to emphasize nearby points, while in sparser areas it is moderately broadened to prevent bias arising from insufficient neighbors. The gravitational kernel $1/d_{ij}^2 + \varepsilon$ offers an inverse-square law-based local enhancement mechanism, emphasizing geometric adjacency at extremely short ranges while maintaining numerical stability. The product of these three factors gives rise to a dual-weighted gravitational force: it leverages density consistency to discern the true structure, employs adaptive distance to regulate the scope of influence, and further enhances neighborhood effects through a gravitational kernel. This approach selectively enhances the structural points under non-uniform sampling and complex noise conditions, effectively suppresses distant and sparsely distributed outliers, and achieves a balance between boundary fidelity and computational robustness.

After calculating the weighted gravitational score for each point, the top $\lambda|P_2|$ points with the highest weighted gravitational scores are retained using the Top $\lambda$ ratio within the leaf, yielding the output $P_t^{(3)}, t=1,2,\ldots,n$ for that sub-leaf. Finally, merge the points within each sub-point cloud to produce the denoised result for the entire point cloud.

$$P_{denoise} = \bigcup_t P_t^{(3)}, t=1,2,\ldots,n \quad (22)$$

# 4 Experiment and result analysis

This section presents experimental validation and result analysis of the proposed adaptive dual-weight gravitational denoising method. First, we introduce the data preparation and evaluation metrics used to measure denoising quality and computational overhead. Subsequently, controlled experiments were conducted on public datasets, comparing different noise ratios and types to validate the method's effectiveness in complex noise environments. Next, we tested the method on real-world data, including publicly available real-world datasets and laboratory-collected point clouds, to evaluate its applicability and robustness in practical scenarios. Finally, through key parameter and structural ablation experiments, we systematically assessed the contributions and synergistic effects of critical modules—adaptive voxel outlier removal, kNN low-density point removal, and dual-weight gravitational scoring—on denoising accuracy and computational efficiency.

## 4.1 Data Preparation and Evaluation Metrics

In this study, the Stanford 3D Scanning Repository is adopted as the primary experimental dataset. This dataset is characterized by high geometric fidelity and rich surface details, while the original models contain negligible noise and can therefore be regarded as quasi-ground-truth references, providing a reliable baseline for evaluation. On this basis, we synthetically inject different types and levels of noise under controlled conditions, enabling a systematic assessment of the proposed method in diverse scenarios in terms of both denoising quality and computational efficiency, and allowing for a precise evaluation of its overall denoising performance [27-29].

According to the noise classification defined in Section 3.1, we contaminated the Bunny model from the Stanford 3D Scanning Repository with three levels of synthetic noise: 5% random noise, 10% random noise, and 10% random noise plus an additional 3% dense random noise. These settings are used to emulate invalid-point noise and anomalous-point noise that may arise under different LiDAR operating conditions. The proposed method is compared against three representative denoising approaches, namely Gravitational function[25], Non-iterative[30], and Local Density and Global Statistic[31]. Fig. 3 illustrates the point cloud models before and after noise injection. As the noise ratio increases, the number of invalid points in the point cloud grows accordingly.

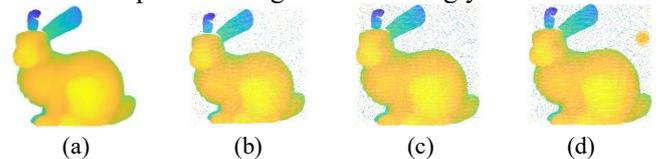

(a)　　　(b)　　　(c)　　　(d)

Fig.3 Original point cloud data and point cloud data after adding random noise. (a) Bunny original point cloud data; (b) Bunny point cloud data with 5% random noise added; (c) Bunny point cloud data with 10% random noise added; (d) Bunny point cloud data with 10% random noise and 3% dense noise added

Table 1 Summary of the number of point clouds before and after noise adding

| Model | Size of point clouds | | | |
|---|---|---|---|---|
| | Original | 5% random noise added | 10% random noise added | 10% random noise and 3% dense noise added |
| Bunny | 40256 | 42269 | 44282 | 45490 |

To further enrich the experiments, point-cloud data acquired by a LARK long-range automotive RUBY PLUS LiDAR (LightIC, Beijing, China) are used to comprehensively validate the effectiveness of the proposed method. In Acquisition Scenario 1, the detection range is 50 m, while in Acquisition Scenario 2 the range is extended to 100 m; the two scenarios differ in spatial scale and noise level, enabling a thorough evaluation of the denoising capability of the proposed algorithm. The data acquisition setups are illustrated in Fig. 4.

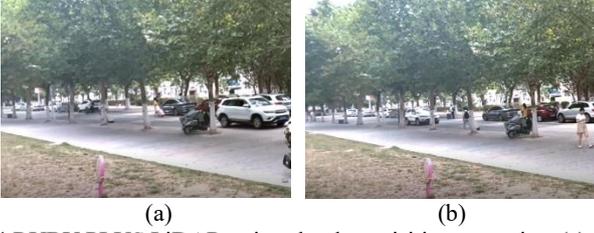

(a) (b)

Fig.4 RUBY PLUS LiDAR point-cloud acquisition scenarios. (a) 50 m data acquisition scenario; (b) 100 m data acquisition scenario.

The ground-truth noise-free point clouds are obtained via manual annotation. The raw acquisition results and the corresponding annotated point clouds are shown in Fig. 5.

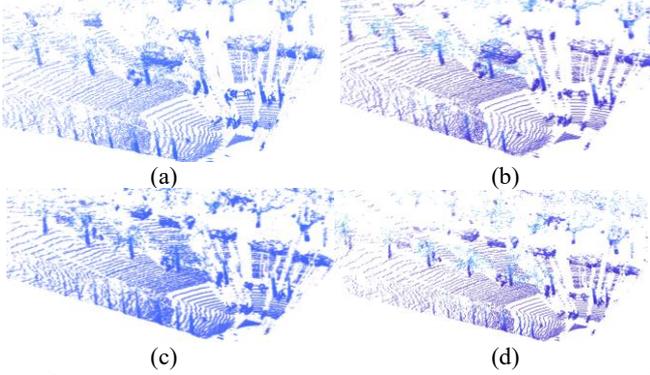

(a) (b)
(c) (d)

Fig.5 Data acquisition results and manually annotated ground truth. (a) and (b) are the raw and annotated point clouds for the 50 m scenario, respectively; (c) and (d) are the raw and annotated point clouds for the 100 m scenario, respectively.

Cohen's Kappa (denoted as κ) is a statistical measure used to quantify the consistency of annotation results, and is commonly applied to assess inter-rater agreement in classification tasks. Unlike directly computing the raw agreement rate, the Kappa coefficient corrects for agreement that may occur by chance, thereby providing a more robust reflection of the true level of consistency.

To ensure the reliability of the RUBY PLUS noise-free point-cloud ground truth, we defined and strictly followed a rigorous annotation protocol. Three independent LiDAR data experts performed blind labeling on the same subset of point clouds: an experienced expert first produced an initial annotation, which was then jointly reviewed and refined by the other two experts until all ambiguous points were resolved. The final ground truth was determined based on the unanimous agreement of all three annotators. The statistics are as follows: in the 50 m scenario, a total of 132,629 points were labeled (63,047 non-noise points and 69,582 noise points); in the 100 m scenario, a total of 133,896 points were labeled (63,737 non-noise points and 70,159 noise points).

If two annotators consistently label a point as "signal", the number of such points is denoted as $N_{agree}^{signal}$; if both label it as "noise", the number is denoted as $N_{agree}^{noise}$. Cases in which their judgments disagree are counted as $N_{disagree}$. The resulting observed agreement ratio is denoted by $p_0^c$, whose computation is given by:

$$p_0^c = \frac{N_{agree}^{signal} + N_{agree}^{noise}}{N_{agree}^{signal} + N_{agree}^{noise} + N_{disagree}} \quad (23)$$

For annotator A, the proportion of points labeled as "signal" is denoted by $p_A^{signal} = N_A^{signal} / N_{total}$, and the corresponding proportion for annotator B is $p_B^{signal} = N_B^{signal} / N_{total}$. Similarly, the proportions of points labeled as "noise" by the two annotators are denoted by $p_A^{noise}$ and $p_B^{noise}$. These probabilities are then used to compute the expected agreement $p_e^c$, as given in Equations (24).

$$p_e^c = p_A^{signal} \cdot p_B^{signal} + p_A^{noise} \cdot p_B^{noise} \quad (24)$$

The resulting Kappa coefficient $\kappa$ is given by Equation (25).

$$\kappa^c = \frac{p_0^c - p_e^c}{1 - p_e^c} \quad (25)$$

After aggregating the annotations from raters A, B, and C and computing the statistics according to Equations (23)–(25), the obtained results are summarized in the table below.

Table 2 RUBY PLUS ground truth results of multi-person annotation

| Data | Pairwise | Consistent number of non noise points | Number of consistent noise points | Number of points of disagreement | $p_0^c$ | $p_e^c$ | $\kappa^c$ |
|---|---|---|---|---|---|---|---|
| 50m data | A vs B | 61257 | 67792 | 3580 | 0.973 | 0.501 | 0.947 |
| | B vs C | 61190 | 67725 | 3714 | 0.972 | 0.501 | 0.945 |
| | A vs C | 61058 | 67593 | 3978 | 0.970 | 0.501 | 0.942 |
| 100m data | A vs B | 61863 | 68285 | 3748 | 0.972 | 0.501 | 0.945 |
| | B vs C | 61796 | 68218 | 3882 | 0.971 | 0.501 | 0.943 |
| | A vs C | 61661 | 68083 | 4152 | 0.969 | 0.501 | 0.938 |

As shown in Table 2, the average $\kappa$ value over all annotator pairs reaches 0.943. A Kappa coefficient in the range of 0.81–1.00 is generally interpreted as indicating almost perfect agreement. Therefore, the manually constructed "noise-free point cloud" ground truth in this study can be regarded as highly accurate and reliable.

In summary, based on the original bunny model from the Stanford 3D Scanning Repository, four frames are constructed:

one noise-free reference frame; two frames with additive random noise at levels of 5% and 10%; and one composite-noise frame containing 10% random noise plus 3% high-density uniform noise. In the RUBY PLUS LiDAR dataset, two additional samples are selected, corresponding to observation scenes at 50 m and 100 m, respectively.

This study adopts a controlled variable experimental design. First, we compare the denoising performance of the proposed method under different noise levels and types to validate its effectiveness. Multiple evaluation metrics are employed, including precision, recall, F1 score, peak signal-to-noise ratio (PSNR), Chamfer Distance (CD), and runtime. Precision measures how accurately noise points are removed, while recall reflects the proportion of true noise points that are successfully eliminated. The F1 score, defined as the harmonic mean of precision and recall, characterizes the overall performance of the noise-suppression method. PSNR is a commonly used indicator for assessing the quality of point-cloud denoising; it quantifies the discrepancy between the denoised point cloud and the original one, with higher PSNR indicating that the denoised result is closer to the ground truth and that noise has been more effectively removed. CD measures the shape discrepancy between the denoised point cloud and the original ground truth, i.e., the degree of geometric consistency between them. Unlike PSNR, CD focuses on geometric structure: a lower CD value indicates that the denoised point cloud more closely matches the ground truth in spatial distribution, implying that the method removes noise while better preserving the geometric details of the object. Conversely, a larger CD indicates more severe loss of true structural details and shape distortion caused by denoising. Finally, runtime is used to evaluate the computational efficiency of each method.

$$P_d = \frac{N_q}{N_s} \quad (26)$$

$$R_d = \frac{N_q}{N_y} \quad (27)$$

$$F1 = \frac{2P_d R_d}{P_d + R_d} \quad (28)$$

where $N_q$ denotes the number of removed noise points, $N_s$ the total number of removed points, and $N_y$ the total number of noise points. The PSNR and CD metrics are computed as follows:

$$PSNR = 10\log_{10}\left(\frac{M^2}{MSE}\right) = 10\log_{10}\left(\frac{M^2}{\frac{1}{n}\sum_{i=1}^n \|p_i^{clean} - p_i^{denoised}\|_2^2}\right) \quad (29)$$

$$CD(A,B) = \frac{1}{|A|}\sum_{p \in A}\min_{q \in B}\|p-q\|_2^2 + \frac{1}{|B|}\sum_{q \in B}\min_{p \in A}\|q-p\|_2^2 \quad (30)$$

In Equations (29), $M$ denotes the maximum distance between any two points in the point cloud, which is usually taken as the length of the diagonal of the bounding box. MSE is the mean squared error, obtained by averaging the squared Euclidean distance between each pair of corresponding points in the ground-truth point cloud and the denoised point cloud. $p_i^{clean}$ is the $i$-th point in the ground-truth cloud, and $p_i^{denoised}$ is its corresponding point in the denoised cloud, typically chosen as the nearest neighbour of $p_i^{clean}$. $\|\cdot\|_2^2$ denotes the squared Euclidean norm. In Equations (30), $A$ and $B$ denote the ground-truth point set and the denoised point set, respectively, and $|A|$ and $|B|$ are the numbers of points in each set. $p$ and $q$ are individual points belonging to sets $A$ and $B$. The operator $\min_{p \in A}$ indicates searching in set $A$ for the nearest neighbour of $q$; the other term is defined in the same way.

### 4.2 Comparative Experiment

To ensure a fair and unbiased comparison, all methods are evaluated under the same hardware environment—AMD Ryzen 7 4800H CPU, NVIDIA GeForce RTX 2060 GPU, 16 GB DDR4 RAM, 512 GB NVMe SSD—and implemented in MATLAB R2023b using exactly the same datasets.

#### 4.2.1 Point cloud denoising experiments under different types of noise

To evaluate the performance of the proposed method under different noise intensities, this subsection designs point cloud denoising experiments at multiple noise levels. Random noise with different ratios is added to the above public dataset, where the noise levels are set to 5% and 10%, respectively.

**(1) Experimental results of adding 5% random noise to point cloud denoising**

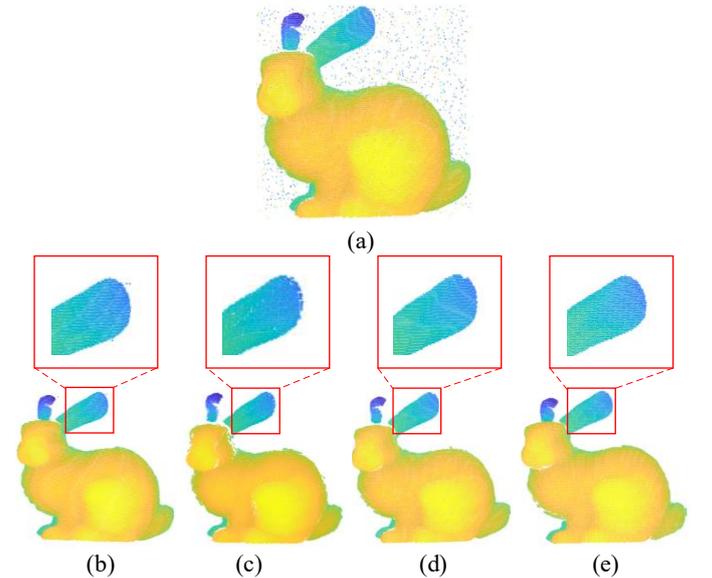

Fig.6 Bunny's denoising results under 5% random noise conditions using various algorithms. (a) Bunny point cloud data with 5% random noise added; (b) Gravitational function point cloud denoising results; (c) Non-iterative point cloud denoising results; (d) Local Density and Global Statistical point cloud denoising results; (e) Results of point cloud denoising algorithm in this article

Table 3 Experimental results of denoising the bunny point cloud under 5% random noise conditions

| Method | Precision | Recall | F1 | PNSR/dB | CD/$mm^2$ | Runtime/s |
|---|---|---|---|---|---|---|
| Gravitational function [25] | 0.9943 | 0.9817 | 0.9880 | 73.8629 | 0.1610 | 0.316 |
| Non-iterative [30] | 0.4716 | 0.9913 | 0.6391 | 37.5118 | 11.1014 | 0.630 |
| Local Density and Global | 0.9888 | 0.9888 | 0.9888 | 74.7945 | 0.2260 | 0.137 |
| Statistic [31] | | | | | | |
| Proposed | 0.9893 | 0.9944 | 0.9918 | 75.6970 | 0.1303 | 0.067 |

**（2）Experimental results of adding 10% random noise to point cloud denoising**

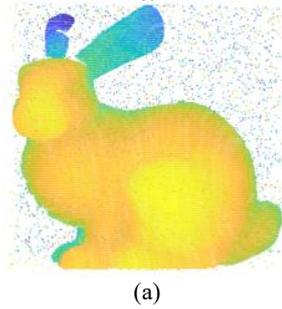

(a)

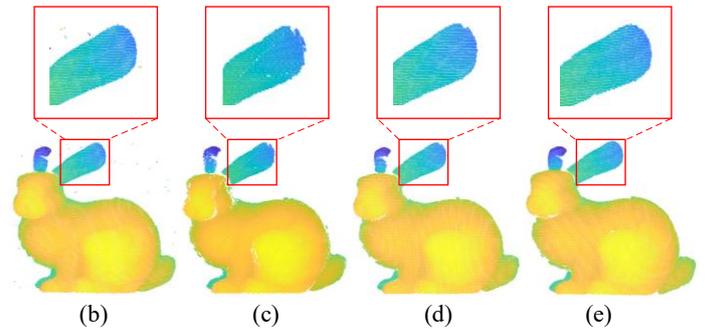

(b) (c) (d) (e)

Fig.7 Bunny's denoising results under 10% random noise conditions using various algorithms. (a) Bunny point cloud data with 10% random noise added; (b) Gravitational function point cloud denoising results; (c) Non-iterative point cloud denoising results; (d) Local Density and Global Statistical point cloud denoising results; (e) Results of point cloud denoising algorithm in this article

Table 4 Experimental results of denoising the bunny point cloud under 10% random noise conditions

| Method | Precision | Recall | F1 | PNSR/dB | CD/$mm^2$ | Runtime/s |
|---|---|---|---|---|---|---|
| Gravitational function [25] | 0.9973 | 0.9529 | 0.9746 | 70.9044 | 1.9727 | 0.310 |
| Non-iterative [30] | 0.6114 | 0.9949 | 0.7574 | 36.7687 | 13.1460 | 0.697 |
| Local Density and Global Statistic [31] | 0.9957 | 0.9949 | 0.9953 | 72.3195 | 0.2131 | 0.144 |
| Proposed | 0.9969 | 0.9952 | 0.9960 | 73.8966 | 0.1901 | 0.075 |

The comparative experiments under different noise levels show that the proposed method exhibits a clear advantage in computational efficiency: its average processing time is consistently the shortest, demonstrating a stronger capability for large-scale point cloud processing. In terms of denoising quality, the proposed method achieves overall higher F1 scores and PSNR than the three baseline methods, indicating superior denoising accuracy. Meanwhile, the CD is consistently the lowest, and the visualization results further confirm that the proposed method better preserves the geometric edges and structural details of the object. These performance gains mainly stem from two design choices: (i) voxel-based gating combined with kNN density thresholds is used to remove clearly isolated points and low-density points in advance, which significantly reduces unnecessary gravitational computations on invalid points; (ii) the gravitational score jointly weighted by local density and adaptive distance assigns larger contributions to points within dense, short-range structures, thereby effectively suppressing noise. Overall, the proposed method achieves the best trade-off between accuracy and efficiency: it attains the shortest denoising time, the highest F1 and PSNR, and the lowest CD, and exhibits stable and reliable denoising performance under 5% and 10% random noise levels.

In addition, we injected random noise with intensities ranging from 15% to 50% to further investigate the impact of noise levels on the performance of the proposed method. The variations of F1 score, PSNR, CD, and processing time with respect to noise level, for the proposed method and the comparative methods, are shown in the following curves.

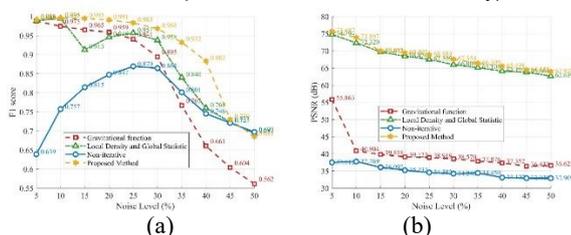

(a) (b)

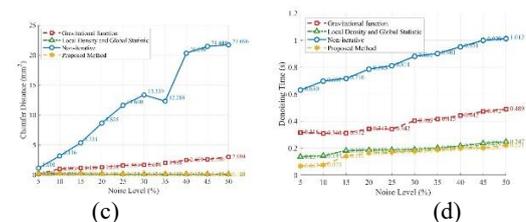

(c) (d)

Fig.8 Curves of the evaluation metrics versus noise level; the yellow curve corresponds to the proposed method

The four sets of curves in Fig. 8 show that the proposed method maintains stable and consistently superior overall performance across different noise levels. In the 5%–45% noise range, the F1 score of the proposed method is clearly higher than that of the other three methods. Although PSNR decreases gradually as the noise level increases, it remains higher than the competing methods, indicating better denoising accuracy. The CD curve increases only slightly with noise level and stays at a relatively low magnitude, which demonstrates that the proposed method is more effective in preserving geometric edge structures. Meanwhile, the denoising time grows only mildly as noise increases, thereby maintaining high computational efficiency.

**4.2.2 Point cloud denoising experiments under different types of noise**

The denoising results for the point cloud with 10% random noise and 3% Gaussian noise are shown in Fig. 9.

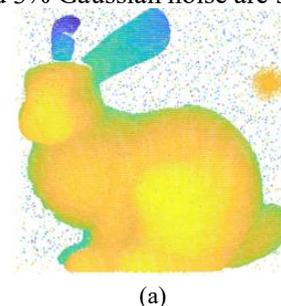

(a)

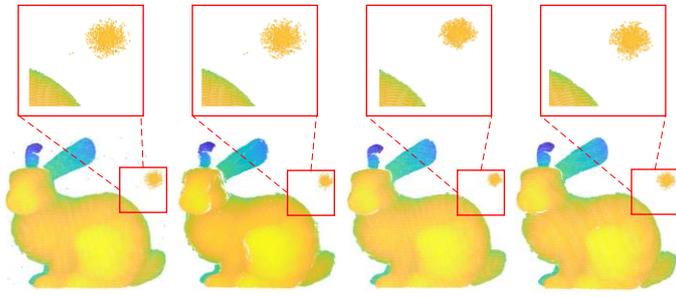

(b)      (c)      (d)      (e)

Fig.9 Bunny's denoising results using various algorithms under 10% random noise and 3% dense noise conditions. (a) Bunny point cloud data with 10% random noise and 3% dense noise added; (b) Gravitational function point cloud denoising results; (c) Non-iterative point cloud denoising results; (d) Local Density and Global Statistical point cloud denoising results; (e) Results of point cloud denoising algorithm in this article

Table 5 Experimental results of denoising bunny point cloud with 10% random noise and 3% dense noise conditions

| Method | Precision | Recall | F1 | PNSR/dB | CD/$mm^2$ | Runtime/s |
|---|---|---|---|---|---|---|
| Gravitational function [25] | 0.9021 | 0.8201 | 0.8591 | 60.1001 | 4.7814 | 5.610 |
| Non-iterative [30] | 0.6547 | 0.8163 | 0.7266 | 41.8742 | 15.1209 | 0.696 |
| Local Density and Global Statistic [31] | 0.9878 | 0.8216 | 0.8971 | 62.8939 | 4.7901 | 0.142 |
| Proposed | 0.9888 | 0.8637 | 0.9220 | 63.7283 | 4.1901 | 0.089 |

In the mixed-noise scenario with 10% random noise and an additional 3% dense noise clusters, the proposed method achieves the best performance across all metrics. The denoising precision and recall reach 0.9888 and 0.8637, respectively, yielding an F1 score of 0.9220. In terms of denoising accuracy, the PSNR attains 63.73 dB, outperforming all three comparison methods. In terms of geometric preservation, the CD of the proposed method is 4.1901 mm², which is lower than the three compared methods. In terms of computational efficiency, the running time is 0.089 s, which is lower than the three comparison methods. Overall, under mixed-noise conditions, the proposed method simultaneously achieves higher denoising accuracy, stronger geometric fidelity, and faster computational speed.

In addition, on top of a fixed 10% random noise level, we further superimpose dense noise clusters with ratios ranging from 8% to 48% in order to evaluate the impact of different dense-noise levels on the proposed method. The variations of F1 score, PSNR, CD, and runtime for the proposed method and the comparison methods as the dense-noise ratio increases are shown in Fig. 10.

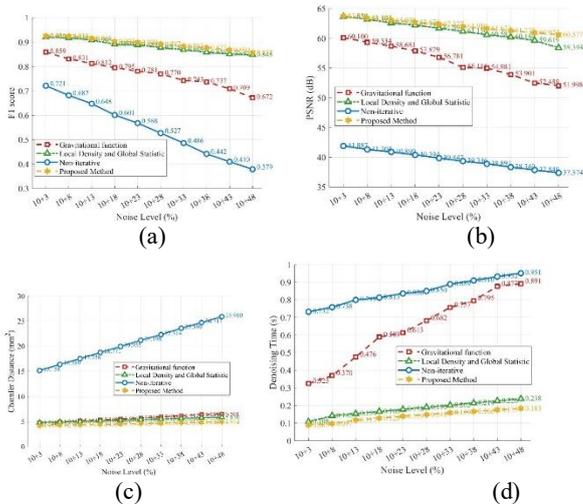

Fig.10 Curves of performance metrics versus dense noise-cluster level

In the presence of 10% random noise combined with varying proportions of dense noise clusters, the proposed method consistently outperforms the competing approaches across all four evaluation metrics. Over the entire noise range, the F1 score remains the highest and decreases only slowly as the noise level increases, clearly exceeding the other three methods. The PSNR curve of the proposed method is also consistently the highest with the smallest degradation, indicating strong robustness under mixed-noise conditions. The maximum CD achieved by our method is only 4.9 mm², which is lower than that of the other methods at the corresponding noise levels, demonstrating superior preservation of geometric boundaries and fine structural details. In addition, the proposed method achieves the shortest runtime and is always the fastest among all methods. Overall, as the mixed-noise intensity increases from 10%+3% to 10%+48%, the proposed approach maintains higher F1 scores, better denoising accuracy and edge preservation, and lower computational cost.

### 4.2.3 Point cloud denoising experiments on real-world public LiDAR datasets

In order to further verify the denoising ability of the proposed method for real-world scene noise, this paper conducted point cloud denoising experiments with SnowNet [32] on the Canadian Adverse Driving Conditions Dataset (CADC) [33], and evaluated it using F1 Score, PSNR, CD, and Runtime metrics. The experimental results are shown in Fig. 11.

The CADC dataset is a public benchmark specifically designed for research on autonomous driving and perceptual systems in challenging environments. It aims to simulate and assess the performance of autonomous vehicles under adverse weather and driving conditions. The dataset, jointly developed by Canadian research institutions and universities, contains driving data acquired in various adverse weather and road scenarios, and is widely used for developing and validating perception, decision-making, and control modules of autonomous vehicles. In the selected point cloud frame used in this study, the original LiDAR scan contains 43,502 points, including 1,495 noise points. The corresponding real-world acquisition scene and point cloud visualization results are shown below.

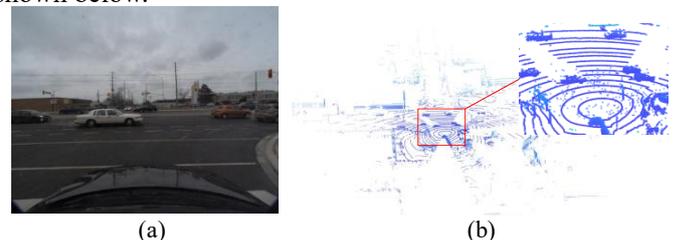

(a)      (b)

Fig.11 Real CADC data acquisition scene and point cloud visualization results. (a) RGB image of the real data acquisition scene; (b) visualization of the acquired LiDAR point cloud.

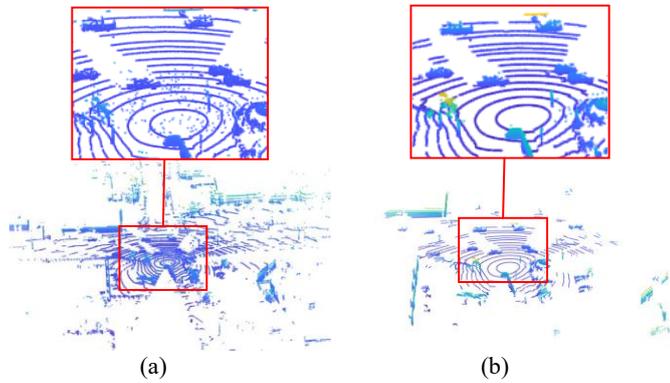

| (a) | (b) |

Fig.12 Denoising results of the proposed method on the CADC dataset. (a) Original noisy point cloud; (b) point cloud after denoising with the proposed method.

Table 6 Point cloud denoising results on the CADC dataset.

| Method | Precision | Recall | F1 | Runtime/s |
|---|---|---|---|---|
| SnowNet[32] | 0.91 | 0.86 | 0.88 | 0.042 |
| Proposed | 0.99 | 0.98 | 0.98 | 0.058 |

From the CADC real-scene experiment, it can be seen that the proposed method maintains stable denoising performance under complex disturbances such as snow particles. Compared with SnowNet, the precision is improved from 0.91 to 0.99, the recall from 0.86 to 0.98, and the F1 score from 0.88 to 0.98, demonstrating a clear overall advantage. Although the runtime reaches 0.058s, it still satisfies near real-time processing requirements. Overall, the proposed method achieves a favorable balance between denoising accuracy and robustness in real snowy scenes, gaining substantial performance improvements at the cost of only a modest increase in computational time.

#### 4.2.4 Point Cloud Denoising Experiments on Laboratory-Collected LiDAR Data

We further employ laboratory-acquired data to evaluate the denoising performance of the proposed method in real-world scenarios. RUBY PLUS LiDAR point clouds are collected at the previously described ranges of 50 m and 100 m, and noise-free ground-truth point clouds are constructed via manual annotation. The experimental results of the proposed method are shown in Fig. 15 and Table 7, respectively.

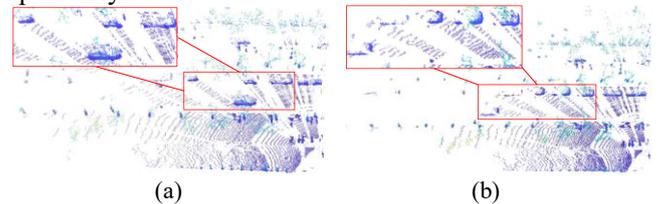

| (a) | (b) |

Fig.15 Laboratory point-cloud denoising results. (a) Denoising experiment for the 50 m object point cloud; (b) denoising experiment for the 100 m object point cloud.

Table 7 Experimental records of point-cloud denoising on laboratory-acquired data

| Distance | Method | Precision | Recall | F1 | PNSR/dB | CD/$mm^2$ | Runtime/s |
|---|---|---|---|---|---|---|---|
| 50m | Local Density and Global Statistic | 0.9913 | 0.7291 | 0.8402 | 72.6870 | 2.5511 | 0.434 |
|  | Proposed | 0.9976 | 0.7318 | 0.8443 | 73.9128 | 2.0291 | 0.396 |
| 100m | Local Density and Global Statistic | 0.9917 | 0.7147 | 0.8307 | 72.1284 | 2.1790 | 0.496 |
|  | Proposed | 0.9989 | 0.7203 | 0.8370 | 73.8901 | 1.6901 | 0.351 |

Since the Non-iterative and Gravitational function denoising methods yield poor results, their metrics are not reported in the table. Experimental results show that, at a distance of 50 m, the proposed method achieves an accuracy, F1 score, and PSNR of 0.9976, 0.8443, and 73.9128 dB, respectively, outperforming the comparison method, while reducing the CD to 2.0291 mm² and the runtime to 0.396 s. At 100 m, the accuracy, F1 score, and PSNR reach 0.9989, 0.8370, and 73.8901 dB, with the CD further decreased to 1.6901 mm² and the runtime reduced to 0.351 s. Overall, these results indicate that the proposed method consistently improves point cloud denoising accuracy and edge preservation at different ranges, while maintaining high computational efficiency.

### 4.3 Ablation experiment

To systematically evaluate the roles and necessity of each component in the proposed method, two groups of ablation studies at the parameter level and structural level are conducted on the bunny model from the Stanford 3D Scanning Repository. For the parameter ablation, with all other configurations fixed, sensitivity analyses are performed on three key hyperparameters: the kNN neighborhood size (K), the density threshold (rho_percentile), and the minimum number of points per voxel (min_vox_count). The variations of F1, PSNR, CD, and runtime under different parameter settings are examined to determine suitable values that balance denoising strength and geometric fidelity. For the structural ablation, we decompose the proposed method into its submodules to evaluate their individual performance and to verify the role each one plays within the proposed method.

#### 4.3.1 Parameter Ablation Experiment

The number of nearest neighbors controls the scale at which density-based outliers are removed. In this experiment, K is set to 4, 8, 12, 16, 20, 24, 28, 32, 36, and 40, while the other two hyperparameters are fixed to their default optimal values: the density threshold(rho_percentile) is 0.2, and the minimum voxel point count(min_vox_count) is 0.4. We evaluate the influence of different K values on the F1 score, PSNR, CD, and runtime. The experimental results are shown in Fig. 16.

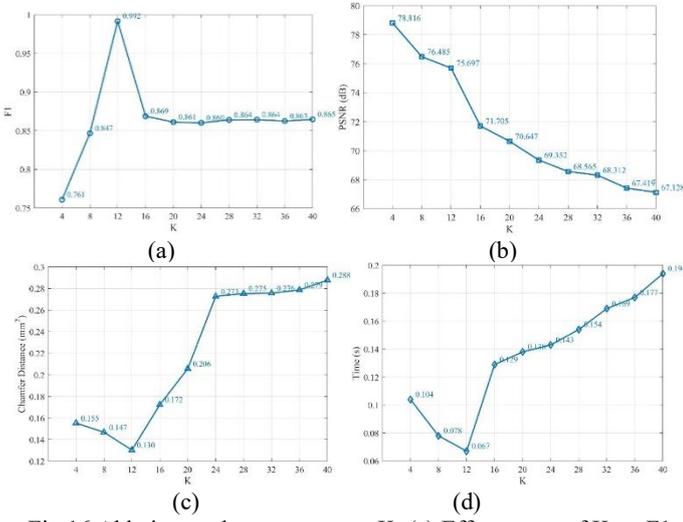

Fig.16 Ablation study on parameter K: (a) Effect curve of K on F1 score, (b) Effect curve of K on PSNR, (c) Effect curve of K on CD, (d) Effect curve of K on runtime.

With all other parameters fixed, the algorithm performance was evaluated under different values of $K$, as shown by the four curves in the figure. In general, when $K$ is small, the neighborhood support is insufficient: although the runtime is shorter, the F1 score is significantly lower. As $K$ increases from 4 to 12, the F1 score improves from 0.76 to 0.99, while PSNR only slightly decreases from 78.8 dB to 75.7 dB and remains at a high level, indicating that $K$ can effectively suppress noise while preserving object points. When $K$ is further increased to 40, the F1 score stabilizes around 0.86 and becomes slightly lower than that at $K$ =12; meanwhile, PSNR decreases monotonically, and CD increases noticeably from its minimum at $K$=12, implying that an excessively large neighborhood leads to over-smoothing of geometry. In addition, the runtime reaches its global minimum at $K$=12, and for $K>12$ it grows approximately linearly due to the increased cost of neighborhood search and gravitational computation. Considering F1, PSNR, CD, and runtime jointly, $K$ =12 can be regarded as a good trade-off between accuracy and efficiency, and is therefore adopted as the default setting in subsequent experiments.

The density threshold (rho_percentile) can determine the intensity of low-density point removal. We set the threshold to 0.05, 0.1, 0.15, 0.2, 0.25, 0.3, 0.35, 0.4, 0.45, and 0.5, respectively. The $K$ value was set to 12, min_vox_count to the default value of 4, and other parameters remained at their default settings. Experiments validated the effects of different density-threshold percentiles on F1 score, PNSR, CD, and runtime. The experimental results are shown in Fig. 17.

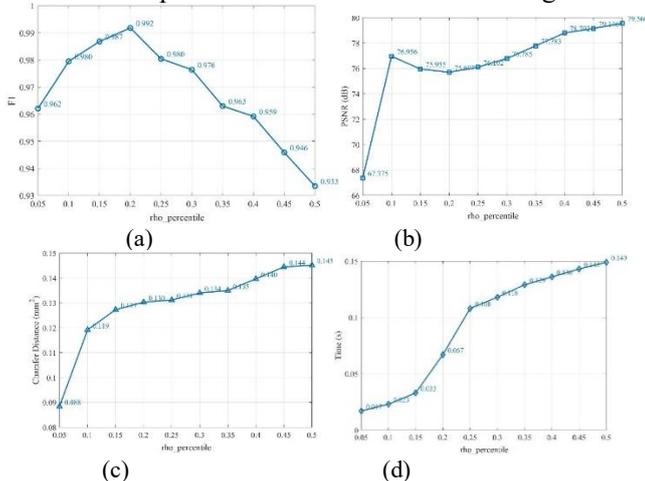

Fig.17 Ablation study on parameter rho_percentile: (a) Effect curve of rho_percentile on F1 score, (b) Effect curve of rho_percentile on PNSR, (c) Effect curve of rho_percentile on CD, (d) Effect curve of rho_percentile on runtime.

As shown in the figure, as rho_percentile increases from 0.05 to 0.2, the F1 score steadily rises from 0.962 to 0.992, indicating that moderately increasing the density threshold helps remove isolated noise points with significantly low local density, thereby improving noise discrimination accuracy. When rho_percentile is further increased to 0.25 and above, the F1 score begins to decline slowly, suggesting that an excessively large threshold leads to the removal of true points in mid-to-long range or locally sparse regions. PSNR increases overall with rho_percentile and tends to saturate after 0.4, reflecting that stricter density constraints can reduce the overall mean squared error; however, the CD value increases monotonically from 0.088 at rho_percentile = 0.05 to 0.145 at rho_percentile = 0.50, indicating that overly strong density suppression introduces geometric over-smoothing and loss of fine details. In terms of runtime, as the threshold increases, the number of points involved in gravitational computation decreases, and the computation becomes more concentrated in high-density regions, resulting in only minor changes in overall processing time. Considering F1, PSNR, CD, and runtime jointly, rho_percentile = 0.2 achieves a favorable balance between suppressing isolated noise and preserving sparse objects, and is therefore adopted as the fixed density threshold in this work.

The min_vox_count can control the intensity of outlier removal. We set the min_vox_count to 2, 3, 4, 5, 6, 7, 8, 9, 10, and 11, respectively, with $K$ set to 12, rho_percentile set to 0.2. The experiments validated the impact of different voxel refinement thresholds on F1 score, PNSR, CD, and runtime. The experimental results are shown in Fig. 18.

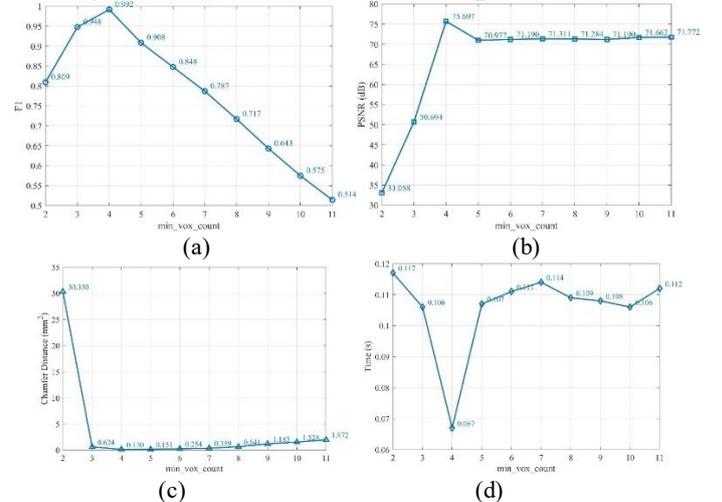

Fig.18 Ablation study on parameter min_vox_count. (a) Effect curve of min_vox_count on F1 score, (b) Effect curve of min_vox_count on PNSR, (c) Effect curve of min_vox_count on CD, (d) Effect curve of min_vox_count on runtime.

As shown in the figure, as min_vox_count increases from 2 to 4, the F1 score rises from 0.809 to 0.992, PSNR increases significantly from 33.06 dB to 75.7 dB, and CD decreases from 30.33 to 0.13, indicating that appropriately increasing the voxel occupancy threshold can effectively remove spatially isolated voxel noise and markedly improve point cloud quality. When min_vox_count is further increased to 5 and above, the F1 score decreases monotonically, and CD continues to grow, suggesting that sparse voxels containing true object points are erroneously removed, thereby damaging the geometric structure of the object point cloud. In terms of runtime, the minimum value in this experiment is achieved at min_vox_count = 4. Considering F1, PSNR, CD, and runtime jointly, min_vox_count = 4 can be regarded as an optimal setting for the voxel gating module and is therefore adopted as the default parameter in the subsequent experiments.

**4.3.2 Structural Ablation Experiment**

To quantitatively evaluate the contribution of each structural module to overall denoising performance, this paper designs ablation experiments at the structural level. This experiment uses the point cloud denoising method based on the Gravitational function described in Section 2.2 as the baseline. Building upon this, we first introduce the octree module to validate its fundamental enhancement to denoising efficiency. Subsequent experiments were conducted on octree-partitioned data, sequentially incorporating the three core modules proposed in this paper: the Adaptive Voxel Outlier Removal module (A1), the kNN Low-Density Point Removal module (A2), and the Weighted Gravitational Value Calculation module (A3). The specific configurations are shown in the table below. By comparing the differences in F1 score, PNSR, CD, and runtime among the models, we aim to quantitatively evaluate the independent contribution and synergistic effect of each module on the final performance.

Table 8 Ablation Study of the Octree Module

| Structure Configuration | F1 | PNSR/dB | CD/$mm^2$ | Runtime/s |
|---|---|---|---|---|
| w/ octree | 0.9877 | 73.4467 | 0.1920 | 0.194 |
| o/ octree | 0.9960 | 73.8966 | 0.1901 | 0.075 |

As shown in the table, the octree-based spatial partitioning module leads to an improvement in overall computational efficiency. Without the octree, the runtime per frame is 0.194s, whereas after introducing the octree, it is reduced to 0.075s. This indicates that octree partitioning refines local neighborhood search and reduces redundant computations without degrading denoising quality.

Table 9 Structural Ablation Experiment Results

| Structure Configuration | A1 | A2 | A3 | F1 | PNSR/dB | CD/$mm^2$ | Runtime/s |
|---|---|---|---|---|---|---|---|
| Baseline | ✗ | ✗ | ✗ | 0.9746 | 70.9044 | 1.9727 | 0.310 |
| OnlyA1 | ✓ | ✗ | ✗ | 0.6822 | 68.0596 | 0.2020 | 0.039 |
| OnlyA2 | ✗ | ✓ | ✗ | 0.5805 | 28.4805 | 86.8933 | 0.045 |
| OnlyA3 | ✗ | ✗ | ✓ | 0.9810 | 71.3328 | 1.0879 | 0.896 |
| A1+A2 | ✓ | ✓ | ✗ | 0.9083 | 69.2271 | 4.2820 | 0.067 |
| A1+A3 | ✓ | ✗ | ✓ | 0.9213 | 71.8031 | 0.4114 | 0.145 |
| A2+A3 | ✗ | ✓ | ✓ | 0.9337 | 73.5248 | 0.8721 | 0.288 |
| Ours | ✓ | ✓ | ✓ | 0.9960 | 73.8966 | 0.1901 | 0.075 |

The structural ablation results show that the individual submodules are highly complementary in terms of accuracy and efficiency. Without any additional modules (Baseline), the method achieves an F1 of 0.9746, a PSNR of 70.9044 dB, and a CD of 1.9727 mm², with a runtime of 0.31 s, while leaving considerable residual noise and incurring a relatively high computational cost. When only A1 or A2 is used, the runtime can be reduced, but the F1 score drops to 0.6822 and 0.5805, respectively, and the PSNR also decreases significantly, indicating that relying solely on voxel or density thresholds tends to mistakenly remove object edge points. When only A3 is used, the F1 score increases to 0.9810, the PSNR to 71.3328 dB, and the CD decreases to 1.0879 mm², demonstrating that the weighted gravitational term helps to finely distinguish objects from noise; however, due to the lack of preliminary filtering, the runtime increases to 0.896 s. After combining modules, A1+A2 can significantly reduce the computational cost, and A1+A3 and A2+A3 further improve accuracy and CD. When all three modules are activated simultaneously (Ours), the F1 rises to 0.996, the PSNR reaches 73.8966 dB, the CD is reduced to 0.1901 mm², and the runtime is only 0.075 s. These results indicate that A1 and A2 are responsible for rapidly discarding most invalid points, while A3 performs fine-grained discrimination on the compressed candidate set, thereby substantially improving overall efficiency while preserving denoising accuracy and edge structures.

# 5 Conclusion

This paper addresses the problem of point cloud denoising and proposes an adaptive dual-weight gravitational-based point cloud denoising method. First, an octree is used to spatially partition the point cloud, enabling local parallelism and scale adaptivity, and providing an efficient data structure for subsequent density estimation and weighted gravitational computation. Then, voxel occupancy counts combined with kNN density estimation are employed to remove most isolated points and low-density noise points, significantly reducing the effective candidate set. Finally, on the candidate points, a dual-weight gravitational scoring function with density weights and adaptive distance weights is applied to perform fine-grained denoising, suppressing various types of noise while preserving edge structures. Experiments on synthetic noisy data from the Stanford 3D Scanning Repository, real snowy scenes from the CADC dataset, and RUBY PLUS LiDAR point clouds acquired in our laboratory demonstrate that the proposed method overall outperforms the comparison methods in terms of F1, PSNR, CD, and runtime, achieving a good balance between denoising accuracy and efficiency.

Although the experimental results show that the proposed method performs well under various noise conditions, it still has limitations in extremely high-noise scenarios and more complex real-world environments. The current experiments mainly cover random noise with a ratio of 5%–50% and mixed cases of "random noise + dense noise clusters". When the noise ratio further increases, the noise clusters become much larger than the object, or the noise is highly similar to the object in terms of local density and scale, the proposed method may mistakenly remove structural points and cause excessive edge smoothing, leading to a decrease in denoising accuracy. In addition, the real-scene validation in this work mainly focuses on snowy weather and highway traffic scenarios, and lacks systematic experimental evaluation under other atmospheric conditions such as rainfall, sandstorms, and dense fog. Future work will focus on multi-scale density modeling and adaptive parameter adjustment, integrating object shape priors and learned features into the gravitational scoring process, and constructing point cloud datasets that cover rain, fog, and sand-related noise, to improve the robustness and engineering applicability of the proposed method in high-noise and multi-weather scenarios.

# Declaration of competing interest

The authors declare that they have no known competing financial interests or personal relationships that could have appeared to influence the work reported in this paper.